\documentclass[10pt,twocolumn,letterpaper]{article}

\usepackage{cvpr_arxiv}
\usepackage{times}
\usepackage{epsfig}
\usepackage{graphicx}
\usepackage{amsmath}
\usepackage{amssymb}
\usepackage{footmisc}
\usepackage{subfig}
\usepackage{stfloats}

\usepackage[pagebackref=true,breaklinks=true,letterpaper=true,colorlinks,bookmarks=false]{hyperref}

\cvprfinalcopy 

\def\cvprPaperID{58} 

\newcommand{\footlabel}[2]{%
    \addtocounter{footnote}{1}%
    \footnotetext[\thefootnote]{%
        \addtocounter{footnote}{-1}%
        \refstepcounter{footnote}\label{#1}%
        #2%
    }%
    $^{\ref{#1}}$%
}

\renewcommand{\footref}[1]{%
    $^{\ref{#1}}$%
}

\usepackage[font={small,it}]{caption}
\ifcvprfinal\fi

\begin{document}

\title{Discovering Style Trends through Deep Visually Aware Latent Item Embeddings\thanks{$^{\tiny{\textcopyright}}$2018 IEEE. Personal use of this material is permitted. Permission from IEEE must be
obtained for all other uses, in any current or future media, including
reprinting/republishing this material for advertising or promotional purposes, creating new
collective works, for resale or redistribution to servers or lists, or reuse of any copyrighted
component of this work in other works}}

\author{ Murium Iqbal, Adair Kovac, Kamelia Aryafar\\
Overstock.com \\
{\tt\small \{miqbal, akovac, karyafar\}@overstock.com}}

\maketitle
\thispagestyle{empty}

\begin{abstract}
 In this paper, we explore Latent Dirichlet Allocation (LDA)~\cite{blei2003latent} and Polylingual Latent Dirichlet Allocation (PolyLDA)~\cite{mimno2009polylingual}, as a means to discover trending styles in Overstock~\footlabel{overstockurl}{\url{www.overstock.com}} from deep visual semantic features transferred from a pretrained convolutional neural network and text-based item attributes. To utilize deep visual semantic features in conjunction with LDA, we develop a method for creating a bag of words representation of unrolled image vectors. By viewing the channels within the convolutional layers of a {\tt Resnet-50}~\cite{he2016deep} as being representative of a word, we can index these activations to create visual documents.  We then train LDA over these documents to discover the latent style in the images. We also incorporate text-based data with PolyLDA, where each representation is viewed as an independent language attempting to describe the same style. The resulting topics are shown to be excellent indicators of visual style across our platform.
\end{abstract}

\section{Introduction}
Overstock~\footref{overstockurl} is an online retailer with the goal of creating \textit{dream homes for all}. Online shoppers browse Overstock's catalog with two distinct goals in mind: (i) finding specific items for specific needs, budget constraints and style preferences or (ii) discovering inspirational styles and new items that complement their existing collection, aesthetics preferences and color palettes. Discovering the underlying style trends can help with both discovery of relevant items and inspirational finds.\\
Online shopping in categories such as fashion, jewelry and furniture is a predominantly visual experience. While style discovery in e-commerce is certainly not a new problem~\cite{hu2014style}, the existing methods primarily rely on topic modeling via LDA for text-based information or implicit user feedback. PolyLDA has recently been used to capture style-coherent embeddings based on visual attributes alone~\cite{hsiao2017learning} though in substantially different ways than our method, to the best of our knowledge. In this paper, we explore style discovery on Overstock~\footref{overstockurl} and propose a multimodal topic modeling approach to infer style from deep visual semantic features transferred from a pretrained convolutional neural network in addition to text-based item attributes.

\section{Style Embeddings}
\label{sec:method}
The process for creating multimodal style embeddings is composed of three main steps: (i) image representations in a bag of visual words format, (ii) text-based item attributes representation and (iii) topic modeling via Mallet's LDA and PolyLDA.

\subsection{Image Representation}
\label{Image Representation}
\paragraph{Layer Selection}
Instead of using traditional transfer learning, we aim to transfer the learned parameters contained within the filters of the convolutional layers (noted as layers from now on) of a pre-trained Resnet ~\cite{rafegas2017understanding}. Each channel represents the response from convolving a learned filter along an input image in horizontal and vertical steps. Once trained, the filters respond to specific patterns, some of which can be interpreted as shapes ~\cite{yosinski2015understanding,zeiler2014visualizing}. By viewing these filters as words and the activations in the channels as indicating presence of these words, we can create a bag of visual words representation of each image. To tabulate which channels are considered active, we take the absolute value of the output directly from a convolutional layer, before the ReLU. We assume if any values within the response grid of the channel exceed a threshold, $t_1$, that this channel is active for the image. Active channels are indexed and these indices are concatenated together to create the image's visual document.

\begin{figure}[htp]
\centering
        \includegraphics[scale=0.5,trim=.5cm .5cm .5cm .5cm,clip]{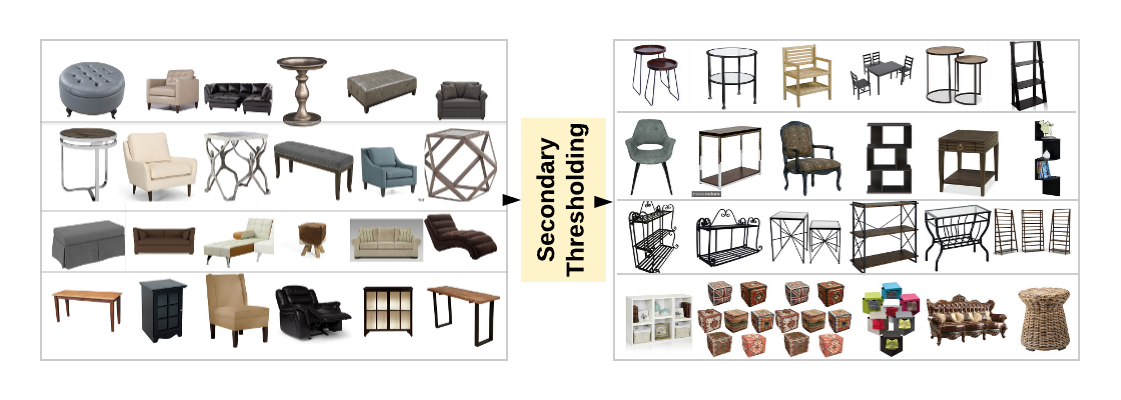}
\caption{Effects of Thresholding on layer $\mathbf{18}$ (a dense layer): There is a distinct, visible improvement on the topics after the thresholding has been applied. Prior to thresholding, it is hard to determine a visual style from the topics. After thresholding, some topics can be seen to contain items with similar shapes or patterns.}
\label{fig:Thresholding}
\end{figure}

\paragraph{Thresholding Activations}
This process for document creation is sensitive to the layer(s). Early layers offer channels with small field of focus that are not well generalized and fire in response to most input images. This results in verbose, redundant documents. Later layers may be too well tuned to the original task of the Resnet and result in sparse uninformative documents for our task. We avoid using later layers, but address the issues of verbose documents by applying a secondary threshold, which is only applied to layers with dense responses. We define dense layers as those which, averaged over a sample set of input images, have non-zero values on more than 1/3 of available channels. These channels are then only considered active if they exceed value $t_1$ on at least 1/20th the grid size of the channel. Applying a secondary thresholding works well for early layers, as can be seen in the resulting topics in figure ~\ref{fig:Thresholding}.  

\begin{figure}[htp]
\centering
        \includegraphics[scale=0.36,trim=.5cm .5cm .5cm .5cm,clip]{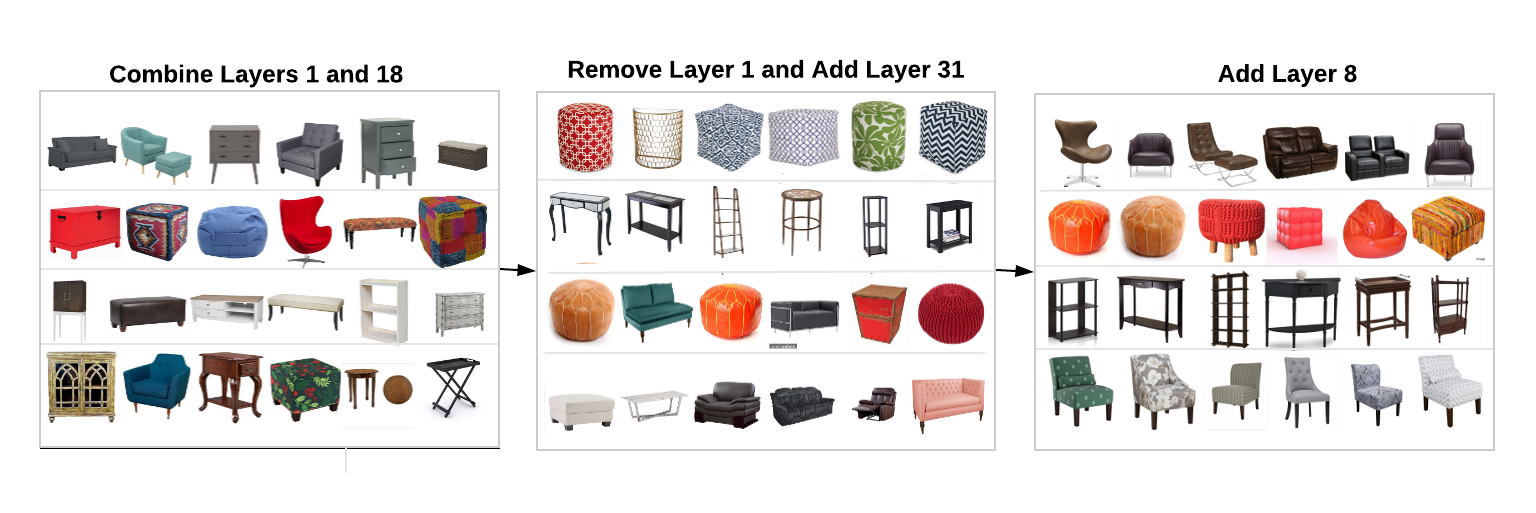}
\caption{Results improve when layers are combined to create a more diverse image based vocabulary. We show results from modeling documents which are a combination of outputs from multiple convolutional layers. Distinct visual styles become increasingly apparent as layers are added. The combination of 3 intermediate layers provides the best results, providing topics that have a cohesive color, furniture style, or material.}
\label{fig:Combining_Layers}
\end{figure}

\paragraph{Combining Layers}
After selecting several layers, we can concatenate the documents from multiple layers together into one document for topic modeling. Based on our results, as depicted in Figure~\ref{fig:Combining_Layers}, combining layers results in better stylized topics. This is likely due to the fact that different layers have different fields of focus on the input, and therefore are tuned to respond to different types, shapes, and sizes of objects. By combining multiple layers we offer the topic modeling a richer visual vocabulary over which to discover the topics.

\subsection{Text Representation}
\label{text}
\begin{figure*}[t]
\centering
\includegraphics[scale=0.725,trim=.5cm .5cm .5cm .5cm,clip]{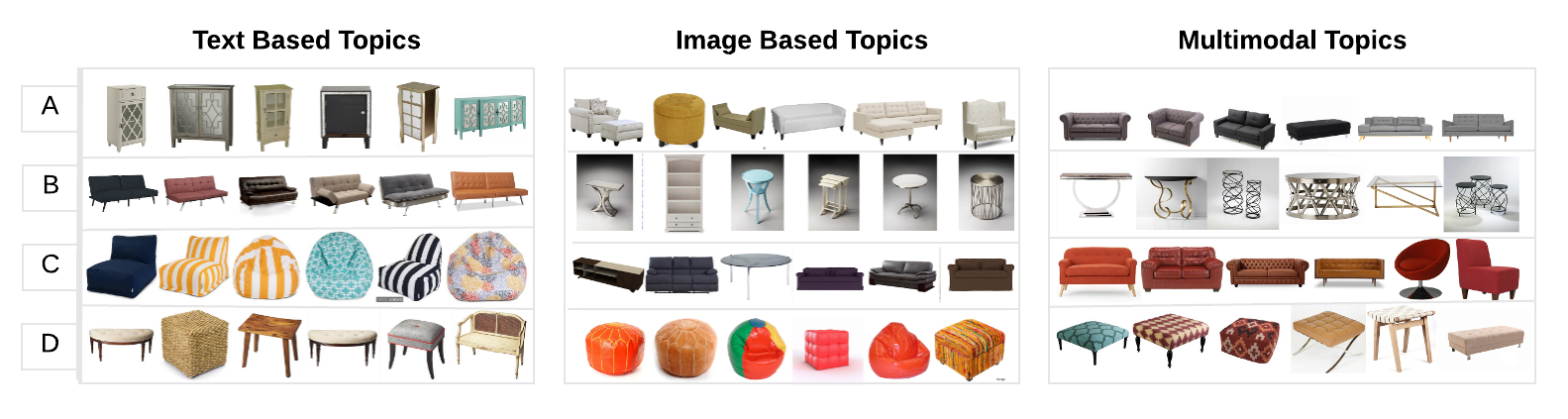}
\caption{Single modality topics generated from LDA on items' text and image (using layers $\mathbf{8}$, $\mathbf{18}$ and $\mathbf{31}$) separately beside multimodal topics generated by PolyLDA. The multimodal model has a topic space that shows the most distinct distributions between inter-pair distances between top-recs and bottom-recs}
\label{fig:Multimodal_Topics}
\end{figure*}

We compare our learned image-based topics to topics generated on text only representations. The text data used for our experiments is a simple bag-of-words representation of item attributes and titles, stripped of stopwords. Attributes are short string descriptors of an item that can help a user navigate to the item. Examples include colors, styles, and materials. Text-only topics also provide good results, but often the topics follow along one category of items. One topic consists of mirrored accent tables, while another consists of futons as seen in Figure~\ref{fig:Multimodal_Topics}.

\subsection{Topic Modeling}
\label{lda}
For discovering the topics, we use LDA, a generative model which assumes the following process for document creation: For a given set of documents, $M$, initialize the set of topic distributions, $\theta_i \sim \mathbf{Dir(\alpha)}$ for \{$i=1\dots $M$\}$. For $K$ topics, initialize the set of word distributions, ${\phi_k} \sim \mathbf{Dir(\beta)}$  for \{$k=1\dots $K$\}$. For the $k^{th}$ word in the $i^{th}$ document select a topic ${z_i}$ from  $\theta_i$ and a word ${w_{i,k}}$ from ${\phi_{z_i}}$ where $\mathbf{Dir}$ is the Dirichlet distribution.

To uncover the styles within our items, we translate images into visual documents by the process described in section ~\ref{Image Representation}. We then assume that the document for an item is the union of all words present in any image associated with the item. The vocabulary depends on which layers we choose to take the channel activations from. These documents are then fed into Mallet's LDA ~\cite{McCallumMALLET}, which performs the topic modeling. The output of the topic modeling provides a representation of each item in the topic space which serve as our style embeddings.

\paragraph{Multimodal Topic Modeling}
We then extend our topic modeling by applying Polylingual LDA by denoting each data modality as different languages. This allows us to incorporate the information available in the text attributes in conjunction with the information from the images. With this assumption we are able to apply the following generative process from PolyLDA: For a given tuple of documents, $\{d_1\dots d_L\}$, initialize a single set of topic distributions, $\theta \sim \mathbf{Dir(\alpha)}$. For $K$ topic sets with $L$ languages, initialize the set of word distributions, ${\phi_{k,l}} \sim \mathbf{Dir(\beta)}$  for $\{k=1\dots K\}$ and $\{l=1\dots L\}$. For the $k^{th}$ word in the $l^{th}$ document in the $i^{th}$ tuple select a topic ${z_{i,l}}$ from  $\theta_i$ and a word ${w_{i,l,k}}$ from ${\phi_{z_{i,l}}}$. The resulting topics can be viewed in Figure ~\ref{fig:Multimodal_Topics}.

\section{Conclusion}
Results from several trained models are provided below, as depictions of high-scoring items from sample topics from our experiments.  We also score the topics for our experiments against user data by taking sets of highly co-clicked items and rarely co-clicked items within our system and seeing how far apart they are within the generated topic spaces. Comparing the topic representations for the same item across the trained models revealed very different representations. Poorer performing topic models had smaller magnitude vectors whose weights were spread across all topics. For the same items, better performing topic models had stronger signals along a few topics and vectors with larger magnitude, clearly associating the items with specific topics. Variants with stronger distributions included those with secondary thresholding, those with multiple Resnet layers, and most clearly the multimodal topics.

We evaluate the models by calculating the inter-pair distance in each topic space of our top 1K pairs of most similar items based on collaborative filtering (CF), and the 1K pairs with the lowest nonzero similarity scores. We will refer to these sets as top-recs and bottom-recs. The text based model provides better results than the image based model, with the text model placing the bottom-recs $3.02\times$ as far apart as the top-recs on average, vs. $1.37\times$ for the image model, although some of this difference may be an artifact of the text but not the image features being useful to find related products on the website when the CF data was collected. Both models showed a roughly Gaussian distribution on the distances for both sets of recs. The multimodal distributions were right-skewed, with the bottom-recs $2.85\times$ further apart than the top-recs on average, but also more heavily skewed than the top-recs so more bottom-recs were further away from the top-recs than in the text-only model. Additionally, since some but not all co-click correlations are generated based on stylistic similarity as perceived by users, the rightward skew itself may indicate better capture of style by this model.


\begin{figure}[htp]
\centering
        \includegraphics[scale=0.28,trim=.5cm .5cm .5cm .5cm,clip]{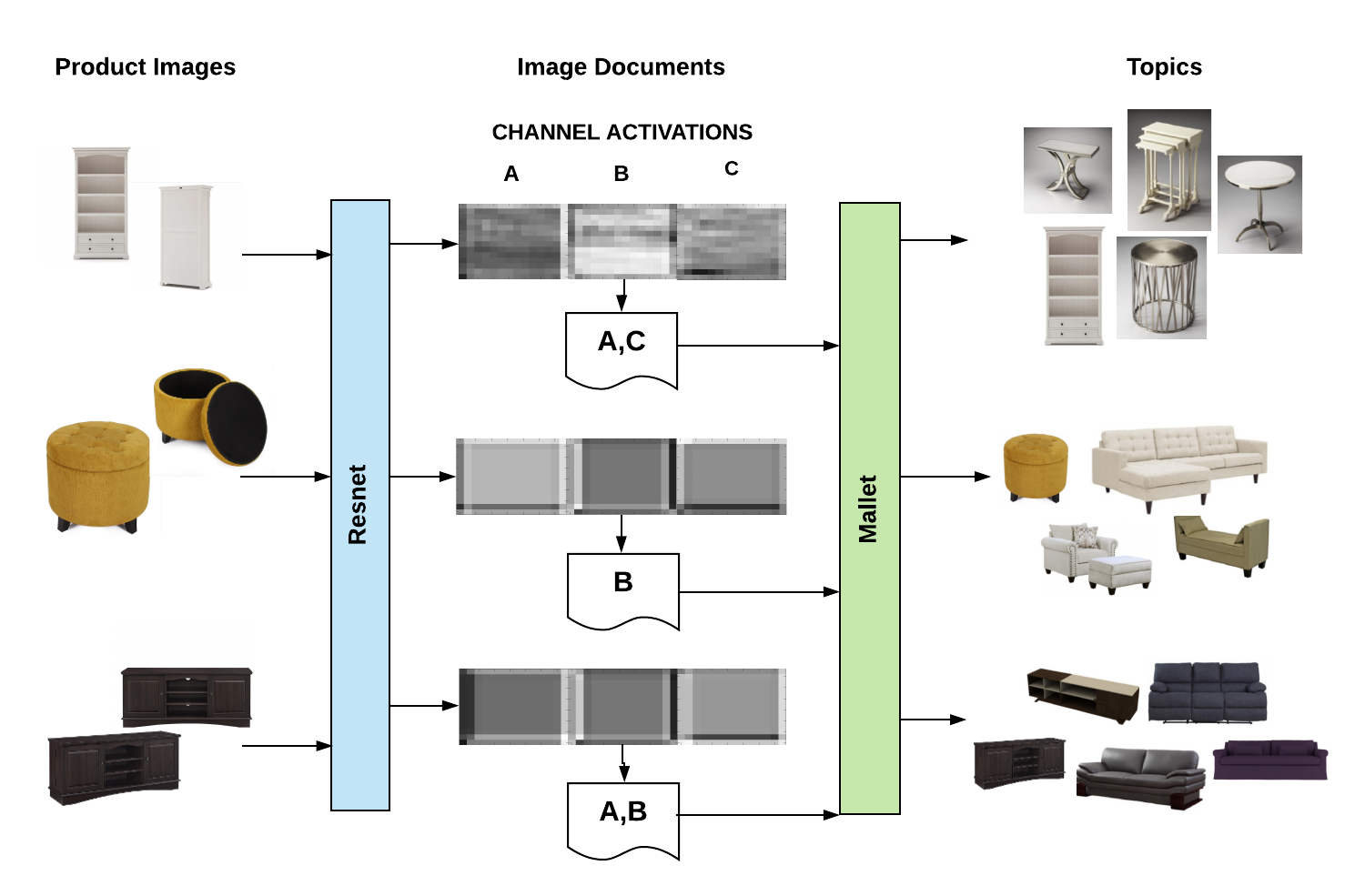}
\caption{Item images are passed through {\tt Resnet-50} to create visual documents. The response from several images are combined to create one document per product. These documents are then passed through Mallet's LDA to discover topics which are indicative of visual styles.}
\vspace{1mm}
\label{fig:Topic_Modeling_Flowchart}
\end{figure}

{\footnotesize
\bibliographystyle{ieee}
\bibliography{egbib}
}

\end{document}